\documentclass{article}

\usepackage{arxiv}

\usepackage[utf8]{inputenc} 
\usepackage[T1]{fontenc}    
\usepackage{hyperref}       
\usepackage{url}            
\usepackage{booktabs}       
\usepackage{amsfonts}       
\usepackage{nicefrac}       
\usepackage{microtype}      
\usepackage{subfigure}

\usepackage{graphicx}
\graphicspath{ {./} }
\usepackage{amsmath}

\usepackage{makeidx}  
\begin{document}
\title{Analysis of hidden feedback loops in continuous machine learning systems}
\renewcommand{\undertitle}{A preprint\textsuperscript{1}}
\renewcommand{\shorttitle}{Analysis of hidden feedback loops in machine learning systems}
\author{Anton S.~Khritankov \\
        Moscow Institute of Physics and Technology \\
        Dolgoprudny, Moscow Region, \\
        Russian Federation \\
        \texttt{anton.khritankov@phystech.edu}}
\date{}

\hypersetup{
pdftitle={Analysis of hidden feedback loops in continuous machine learning systems},
pdfsubject={cs.LG,cs.SE},
pdfauthor={Anton S.~Khritankov},
pdfkeywords={machine learning, continuous machine learning, software quality, feedback loop},
}

\maketitle              

\begin{abstract}        
In this concept paper, we discuss intricacies of specifying and verifying the quality of continuous and lifelong learning artificial intelligence systems as they interact with and influence their environment causing a so-called concept drift. We signify a problem of implicit feedback loops, demonstrate how they intervene with user behavior on an exemplary housing prices prediction system. Based on a preliminary model, we highlight conditions when such feedback loops arise and discuss possible solution approaches.

\end{abstract}

\footnotetext[1]{The final authenticated publication is available online at \url{https://doi.org/10.1007/978-3-030-65854-0_5}}

\section{Introduction}

Definition of quality and requirements specification are important practices in information systems development. While in the software engineering necessary software quality standards (such as ISO 25000 series \cite{iso25000}) have been developed, machine learning systems lack any comparable documents. Being an important part of the system development and operations processes, quality definition and evaluation is a new and active area of research for AI and data-driven systems: autonomous vehicles, self-learning or lifelong machine learning systems \cite{lifelong}, information retrieval and web search systems and others.

\paragraph{Closed loop systems.} In this paper, we consider "closed loop" and partially closed loop machine learning systems, that is, a kind of systems which behavior depends on models learnt from their internal data and inputs provided to the system. An example of such a system could be a recommendation engine for an e-commerce web site or a social network, a digital library search system, job postings search engine and similar digital bulletin systems with content search or recommendation capabilities. In contrast, unless personalized search is enabled, web-scale search engines are not typically considered closed loop (see also \cite{bubble1}) because their response depends on the data out of direct control of the system. Non-closed loop systems also include any systems that do not use data they produce for training.

The problem we study is closely related to the concept drift phenomenon. The so called concept drift is an observable change in input data distribution that occurs over time. For closed loop systems such a drift may occur as a result of changes in the user behavior due to social and cultural reasons, or because of users interaction with the system. The latter effect is called a \textbf{feedback loop}.

\paragraph{Non-observability.} The reasons why discovering these feedback loops important are twofold. First, the effect of a feedback loop is not immediate and may not exhibit itself during testing. Second, in continuous and lifelong machine learning systems algorithms are able to learn from the data present in the system during the lifetime with little or no supervision from developers. These changes are often constitute themselves in internal parameters inside prediction models and may not be directly observable or interpretable. The latter - non-observability - distinguishes feedback loops in machine learning systems from similar loops, for example, in the ordinary software-only or technical systems. What is visible and observable about feedback loops are their effects on user behavior.

\paragraph{Main contributions.} Primary results of this paper are as follows. First, we provide a simple example of a continuous machine learning system with a positive feedback loop, which can be used as a model for further research. Second, based on the model we identify and propose an approach that could be followed to detect feedback loops in simple cases and provide directions for further research.

In the next section we demonstrate a feedback loop effect in an experiment with a housing pricing estimation system. In section 3 we discuss conditions when such feedback loops may occur and how they can be discovered. In section 4 we connect our study with related research in the field.

\section{An exemplary system with a feedback loop}
\subsection{A housing prices prediction problem}
Let us consider a simple and a well-known housing prices prediction problem on the Boston dataset\cite{boston}. Given features $x_i$ of house $i$ predict its price $y' = f(x_i;\theta)$ so that empirical loss on a given dataset $(X, y)$, where $X = \{x_1,..x_n\}, y = \{y_1,.. y_n\}$ is minimized with regards to $\theta$, that is

\[L(y, f(X;\theta)) \rightarrow \underset{\theta}{\text{min}}. \]

In order to evaluate the feedback effect, we use a linear model with high bias and low variance, and a non-linear model with high variance and low bias. The linear model $y' = X \, \theta + b$ is solved as ridge regression with mean squared error loss function $L(y, y')$. The non-linear model is a gradient boosted decision tree regression (GBR) algorithm (both as of scikit-learn implementation \cite{scikit}) with Huber loss function and mean absolute loss (MAE) splitting criterion.

Resembling a typical data science model development process, we employ cross-validation for hyper-parameter tuning for each algorithm and evaluate on held-out data using coefficient of determination $R^2$.

In addition, we use a simple heuristic that helps improve the performance of the models. Notice that price $y$ distribution in the dataset is not symmetric, that is, there can usually be no negative prices, $y \geq 0$. It is known from other domains that relative variation in the price is seen by consumers as more important than absolute change. That is, it is more common to see “market grew $1.5\%$ yesterday”, than "market grew $100$ points". Therefore, we transform $y \leftarrow \log y $.

For the linear model we also transform the source data to zero mean and unit variance before applying the model.

\subsection{Description of the system}

In this experiment, we will run a simulation of a theoretical housing prices prediction system that uses the linear and non-linear models from the previous section. The source code and reproducibility guidelines for the experiment is available on Github\footnote{\url{https://github.com/prog-autom/hidden-demo}}.

When deployed, the system would provide users with an estimate of a house price given its features. Some of the features may be obtained from publicly available sources and depend on house location, while others are specific to the house itself. As both buyers and sellers see the same estimate, they would consider the price as unbiased and treat it as a sort of market average price. This is an important assumption that influences the feedback effect.

In order to model different levels of closedness of the system, we assume that a user either ignores the estimate with probability $1 - p$, either adheres to it with $p$. If adheres, a user chooses a logarithm of the price $\log z_i$ (recall the transform) randomly by sampling it from the Normal distribution $\mathcal{N}(f(x_i; \theta), s \, \sigma_f^2)$, where $\sigma_f^2$ is the model's mean squared error on held-out data and $s > 0$ is a hyperparameter indicating adherence.

\[    \log z_i \sim \begin{cases}
    \mathcal{N} (f(x_i; \theta_, s \, \sigma_f^2), \text{ with probability } p, \\
    \log y_i, \text{ with probability } 1 - p.
    \end{cases}
\]

Because of logarithmic scaling of target variable $y$, actual prices can be calculated by exponentiating the predicted values when needed.

\begin{figure}
    \centering
    \includegraphics[width=0.75 \linewidth]{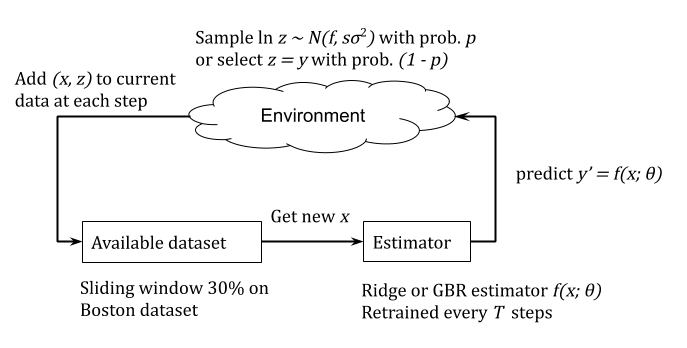}
    \caption{Feedback loop experiment setup}
    \label{fig:mathmodel}
\end{figure}

\subsection{Experiment setup}

The experiment starts with $30\%$ of original data at round $r = 1$. The first model is trained with cross-validation on the $75\%$ part of the starting data giving $\theta^{[r=1]}$ and evaluated on the rest held-out $25\%$ of the data. Then on each step $t \geq 0$ a user takes a prediction $y'_k = f(x_k; \theta^{[r]})$, $k = 0.3n + t$ from the model and decides on the price $z_k$ as specified above. 

The actual price $z_k$ that a user has decided upon and features $x_k$ are appended to the current data and the first item is removed so overall size of current data remains constant. This is equivalent to using a sliding window of the most recent $30\%$ of data with some of original prices $y_k$ in the dataset replaced with user decisions $z_k$.

After each $M$ steps the round increments $r \leftarrow r + 1$ and the model is retrained with cross-validation on current data, which is again split on training $75\%$ and held-out $25\%$ parts giving $\theta^{[r]}$.

The procedure repeats while there are unseen points available $0.3n + t \leq n$. Thus, at each round we know the coefficient of determination $R^2(r)$ and mean absolute error $\text{MAE}(r)$ for each of the models.

\subsection{Results and observations}

\begin{figure}
    \centering
    \begin{subfigure}
        \centering
        \begin{minipage}[t]{0.45\textwidth}
            \includegraphics[width=\textwidth]{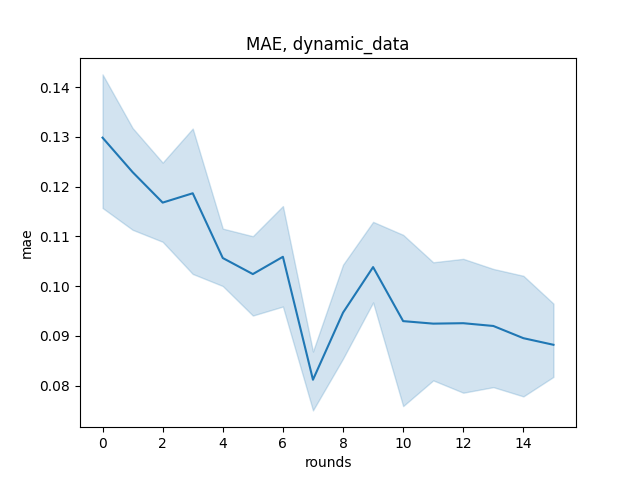} 
            \centering 
            \par{\footnotesize $p=0.7$, $s=0.3$}
        \end{minipage}
    \end{subfigure}
    \begin{subfigure}
        \centering
        \begin{minipage}[t]{0.45\textwidth}
            \includegraphics[width=\textwidth]{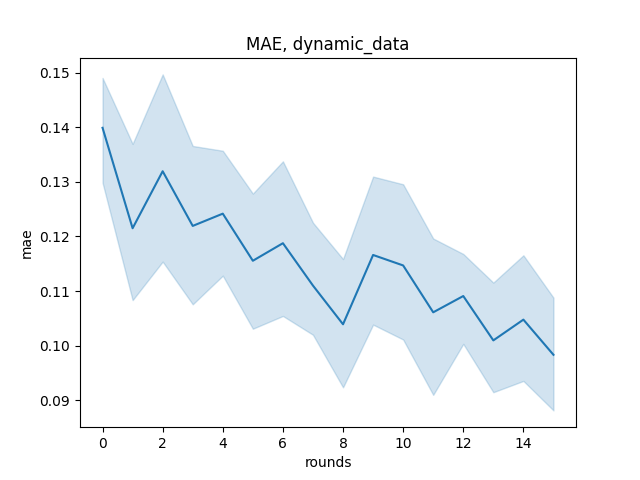}
            \centering 
            \par{\footnotesize $p=0.5$, $s=0.3$}
        \end{minipage}
    \end{subfigure}
    \begin{subfigure}
        \centering
        \begin{minipage}[t]{0.45\textwidth}
            \includegraphics[width=\textwidth]{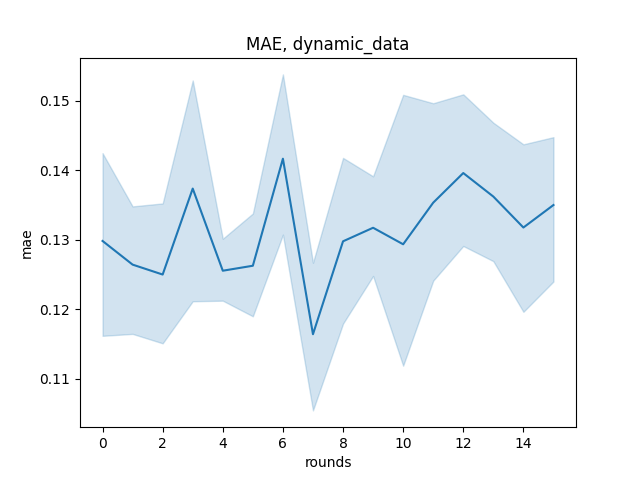}
            \centering 
            \par{\footnotesize $p=0.7$, $s=0.9$}
        \end{minipage}
    \end{subfigure}
    \begin{subfigure}
        \centering
        \begin{minipage}[t]{0.45\textwidth}        
            \includegraphics[width=\textwidth]{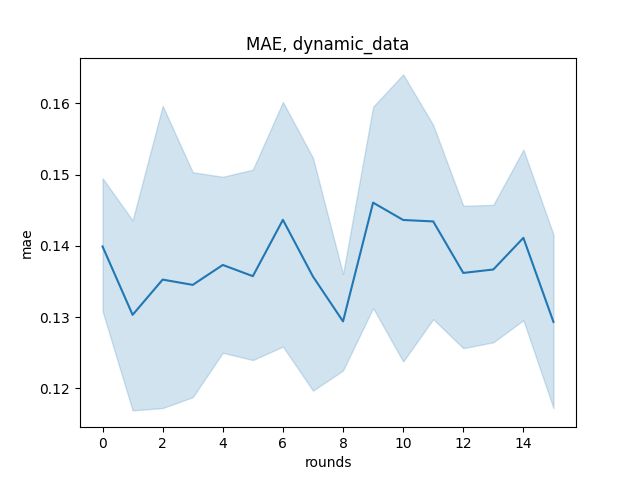}
            \centering 
            \par{\footnotesize $p=0.5$, $s=0.9$}
        \end{minipage}
    \end{subfigure}    
    \caption{Example of a positive feedback loop. Metric: MAE. Model: GBR, steps before retraining $M = 20$. Top left: usage $p = 0.7$, adherence $s = 0.3$, top right: $p = 0.5$, $s = 0.3$. Bottom left: $p = 0.7$, $s = 0.9$, bottom right: $p = 0.5$, $s = 0.9$}
    \label{fig:boosting_results}
\end{figure}

\begin{figure}
    \centering
    \begin{subfigure}
        \centering
        \begin{minipage}[t]{0.45\textwidth}
        \includegraphics[width=\textwidth]{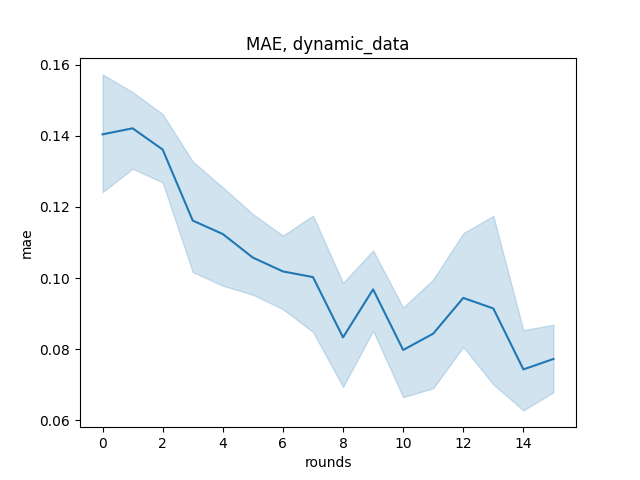}
            \centering 
            \par{\footnotesize $p=0.7$, $s=0.3$}
        \end{minipage}
    \end{subfigure}
    \begin{subfigure}
        \centering
        \begin{minipage}[t]{0.45\textwidth}
        \includegraphics[width=\textwidth]{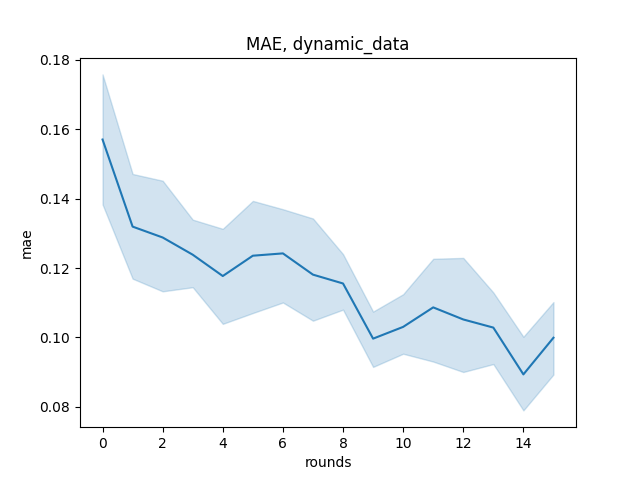}
            \centering 
            \par{\footnotesize $p=0.5$, $s=0.3$}
        \end{minipage}
    \end{subfigure}
    \begin{subfigure}
        \centering
        \begin{minipage}[t]{0.45\textwidth}
        \includegraphics[width=\textwidth]{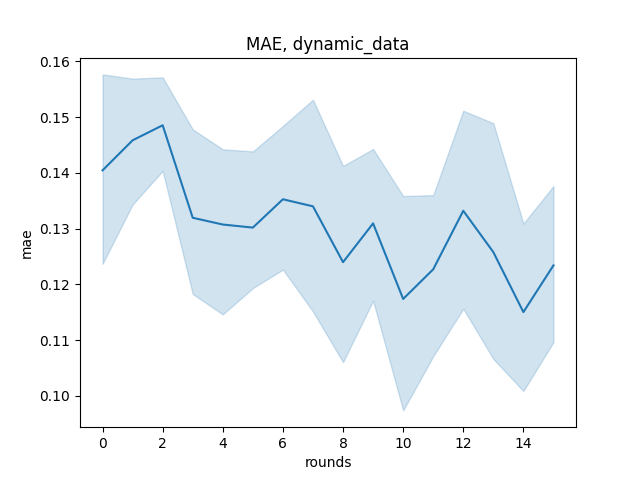}
            \centering 
            \par{\footnotesize $p=0.7$, $s=0.9$}
        \end{minipage}
    \end{subfigure}
    \begin{subfigure}
        \centering
        \begin{minipage}[t]{0.45\textwidth}
        \includegraphics[width=\textwidth]{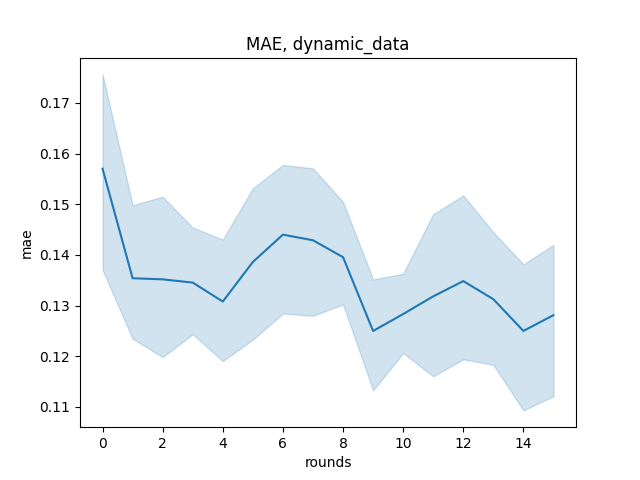}
            \centering 
            \par{\footnotesize $p=0.5$, $s=0.9$}
        \end{minipage}
    \end{subfigure}    
    \caption{Example of a positive feedback loop. Metric: MAE. Model: Ridge, steps before retraining $M = 20$. Top left: usage $p = 0.7$, adherence $s = 0.3$, top right: $p = 0.5$, $s = 0.3$. Bottom left: $p = 0.7$, $s = 0.9$, bottom right: $p = 0.5$, $s = 0.9$}
    \label{fig:ridge_results}
\end{figure}

The representative results are shown at Fig. \ref{fig:boosting_results} and Fig.\ref{fig:ridge_results}. In both cases the model starts getting higher $R^2(r)$ score on held-out data as the number of rounds increases, and tends to $R^2 = 1.0$. Despite the linear regression model having lower quality score in the beginning, it starts outperforming the gradient boosting tree regression (GBR) algorithm after several rounds.

If all users adhere to predictions of the system, that is, the usage parameter $p$ gets closer to $1.0$, the sequence $R^2(r)$ tends to $1.0$ faster. When adherence $p < 0.5$ the $R^2$ may not get to $1.0$ and even decrease as a result of users random sampling of $z_k$, which are added to the current data.

When adherence parameter $s$ is close to $0.0$ and the model has high initial $R^2$ the sequence proceeds faster to $1.0$. Large values of $s > 0.5$ may lead to a lot of noise being added to current data over rounds. Experimentally, $s < 1.25$ at $p = 0.75$ is needed for $R^2(r)$ to converge close to $1.0$ with the linear model, and $s < 0.5$ with the GBR algorithm.

When the number of steps between rounds $T$ gets closer to $1.0$ the $R^2(r)$ of the GBR algorithm becomes noisy and does not grow to $1.0$. Contrary to the linear model, for which $R^2(r)$ tends to $1.0$ for a much wider range of parameters.

For reference, in a completely closed loop system, we would have $p = 1.0$ and $p = 0.0$ for a completely open loop system.

\section{Analysis and discussion}
\subsection{Existence of a positive feedback loop}

It looks like that convergence of the $R^2(r)$ or other performance metrics when round $r \rightarrow \infty$ requires that the deviation of observed data from model predictions should decrease as the number of rounds grows. If it is so, then the deviation would converge to zero and the corresponding model would provide for the minimum possible error.

If we consider a closed-loop lifelong machine learning system as a mapping $T: X \rightarrow X$ on datasets space $X$ (which is the environment in this case), where a single user interaction with the system is an application of the mapping, then we can draft the following.

\emph{Conjecture 1. (existence of positive feedback loop)} A positive feedback loop in a system $T: X \rightarrow X$ exists if
\[
\forall x, y \in X: d(R_f(T(x)), R_f(T(y))) \leq A \cdot d(R_f(x), R_f(y)),
\]
where $0 < A < 1$ is a constant, and $d(r_1,r_2)$ is some metric on model $f$ performance measures $R_f(x)$ over probabilistic space of datasets $X$.

We have not found any similar proven statements for lifelong machine learning systems or feedback loops in particular. However, there are similar results in related fields \cite{nonlinear,fixedpoint}.

\subsection{A checklist for detecting feedback loops}

Following from our findings, we suggest that requirements analysis and review for machine learning systems shall include studies for feedback loops. If such loop is possible, then system requirements should include implementation of measures for detecting and measuring the effect. Therefore, making it observable.

In order to check whether a machine learning system may be affected by feedback loops the following checks may be performed. First, look at the problem statement and where the system receives data from. If the system receives data from the same users that rely on predictions of the model or environment affected by the user behavior, then there is an indication of possible feedback loop.

Second, check the expected impact of the system on user behavior and the data received from users and the environment. If usage $p$ and adherence $s$ parameters may get  $p > 0.5$ and $s < 1.0$ then there is a possibility of feedback loop.

Third, given an indication of a possible feedback loop, we can use definition or Conjecture~1 to check for existence of feedback loops at \emph{test time} or on a simulation model. Set up a baseline algorithm, preferably with low variance and sample pairs of datasets from the environment. If conditions for Conjecture~1 hold then there will be a feedback loop when the system is implemented and  deployed.

Another option is to check for feedback loops at runtime. According to the \emph{No Free Lunch} theorem \cite{nfl}, the performance of the baseline model shall not improve over time. A low variance model with a stable learning algorithm would be useful for this. Alternatively, a range of concept drift detection methods may be used \cite{drift}.

This approach can also be used when a positive feedback loop is a desirable effect to change user behavior or the environment.

\section{Related work}
There is a recent survey on how to adapt machine learning models to concept drift \cite{drift}. Concept drift usually assumes that data distribution is stationary, changes are unexpected and shall be detected and taken into account. In contrast, if feedback loops exist,  input data distribution changes as a result of using the system, therefore changes are expected but may be non-observable.

Prior studies also indicate the importance of feedback loop effects in machine learning systems and relate them to hidden technical debt \cite{hidden} as an unsolved issue. Authors signify that hidden feedback loops shall be identified and removed whenever possible. Monitoring and restricting system actions are also recommended to limit the impact.

In \cite{bing} authors study effects of feedback loops in complex users and advertisers interactions in ads auction in an online advertising system. They notice that temporarily popular ads may get permanent dominance because of positive feedback loops.  Another paper \cite{safety} considers feedback loops in a context of AI systems safety, signifying instabilities and undesired side-effect associated with uncontrolled feedback loops.

In social sciences, echo chambers and filter bubbles are a similar effect produced by feedback loops in content recommendation and search systems \cite{bubble1,bubble2,chamber}. Ensign et al. \cite{runaway} describe a positive feedback loop effect in a predictive policing. Authors consider how the predictive policy system that assigns police patrols influences the city crimes data that is collected back and affects the system itself.

\section{Conclusion and future research}

In this paper we consider a feedback loop problem in closed loop lifelong machine learning systems. We demonstrate and quantify the effect on an exemplary housing prices recommendation system. Based on preliminary findings and analysis of the experiment results, we propose specific measures to check for and detect feedback loops in machine learning systems.

Future research may include formal proof of the conjecture on the existence of feedback loops. Suggestions given in this paper will need to be checked empirically on real-life closed loop systems.

Authors propose to include observability of feedback loops in  quality criteria for machine learning systems interacting with their environment and users. Additional guidelines on satisfying the criteria should be developed as well as supporting software tools.

%
%
\bibliographystyle{plain}

%
\end{document}